\documentclass[sigconf]{acmart}
\AtBeginDocument{%
  }

\setcopyright{acmlicensed}
\copyrightyear{2018}
\acmYear{2018}
\acmDOI{XXXXXXX.XXXXXXX}
\acmISBN{978-1-4503-XXXX-X/2018/06}

\usepackage{amsmath}
\usepackage{amsfonts}
\usepackage{algorithm}
\usepackage[noend]{algorithmic}
\usepackage{graphicx}
\usepackage{textcomp}
\usepackage{xcolor}
\usepackage{enumitem}
\usepackage{booktabs}
\usepackage{multirow}
\usepackage[super]{nth}
\usepackage{tikz}

\usetikzlibrary{tikzmark}
\tikzset{circledColor/.style={circle,draw,inner sep=0.1em,line width=0.04em}}
\usepackage{array}
\newcolumntype{?}{!{\vrule width 1.5pt}}
\usepackage{esvect}
\usepackage{gensymb}
\usepackage{setspace}

\usepackage[flushleft]{threeparttable}
\usepackage{pifont}
\usepackage{caption}

\usepackage{fancyhdr}
\fancypagestyle{firstpage}{
  \fancyhf{}
  \chead{Accepted at the 63rd ACM/IEEE Design Automation Conference (DAC), July 26$^{\text{th}}$ - 29$^{\text{th}}$, 2026 in Long Beach, CA, USA.}
  \cfoot{\scriptsize \thepage}
}

\begin{document}

%%%%%%%%%%%%%%%%%%%%%%%%%%%%%%%%%%%%%%%%%%%%%%%%%%%%%%%%%%%%%%%%%%%%%%%%%%%%%%%%%%
%%%%%%%%%%%%%%%%%%%%%%%%%%%%%%%%%%%%%%%%%%%%%%%%%%%%%%%%%%%%%%%%%%%%%%%%%%%%%%%%%%
%% The "title" command has an optional parameter,
%% allowing the author to define a "short title" to be used in page headers.
\title{Scalable Multi-Task Learning through Spiking Neural Networks with Adaptive Task-Switching Policy for Intelligent Autonomous Agents}

%% The "author" command and its associated commands are used to define
%% the authors and their affiliations.
%% Of note is the shared affiliation of the first two authors, and the
%% "authornote" and "authornotemark" commands
%% used to denote shared contribution to the research.
\author{Rachmad Vidya Wicaksana Putra, Avaneesh Devkota, Muhammad Shafique}
\affiliation{%
  \institution{eBRAIN Lab, New York University (NYU) Abu Dhabi}
  \city{Abu Dhabi}
  \country{United Arab Emirates}
}
\email{{rachmad.putra, ad5768, muhammad.shafique}@nyu.edu}

%%
%% By default, the full list of authors will be used in the page
%% headers. Often, this list is too long, and will overlap
%% other information printed in the page headers. This command allows
%% the author to define a more concise list
%% of authors' names for this purpose.
% \renewcommand{\shortauthors}{Trovato et al.}

%%%%%%%%%%%%%%%%%%%%%%%%%%%%%%%%%%%%%%%%%%%%%%%%%%%%%%%%%%%%%%%%%%%%%%%%%%%%%
%%
%% The abstract is a short summary of the work to be presented in the
%% article.
\begin{spacing}{0.98}
\begin{abstract}
Training resource-constrained autonomous agents on multiple tasks simultaneously is crucial for adapting to diverse real-world environments.
Recent works employ reinforcement learning (RL) approach, but they still suffer from sub-optimal multi-task performance due to \textit{task interference}.
State-of-the-art works employ Spiking Neural Networks (SNNs) to improve RL-based multi-task learning and enable low-power/energy operations through network enhancements and spike-driven data stream processing. 
However, they rely on fixed task-switching intervals during its training, thus  limiting its performance and scalability.
To address this, we propose \textit{\textbf{SwitchMT}}, a novel methodology that employs adaptive task-switching for effective, scalable, and simultaneous multi-task learning. 
SwitchMT employs the following key ideas: (1) leveraging a Deep Spiking Q-Network with active dendrites and dueling structure, that utilizes task-specific context signals to create specialized sub-networks; and (2) devising an adaptive task-switching policy that leverages both rewards and internal dynamics of the network parameters. 
Experimental results demonstrate that SwitchMT achieves competitive scores in multiple Atari games (i.e., Pong: -8.8, Breakout: 5.6, and Enduro: 355.2) and longer game episodes as compared to the state-of-the-art. 
These results also highlight the effectiveness of SwitchMT methodology in addressing task interference without increasing the network complexity, enabling intelligent autonomous agents with scalable multi-task learning capabilities. 
\vspace{-0.3cm}
\end{abstract}

\settopmatter{printacmref=false} % Removes citation information below the abstract
\renewcommand \footnotetextcopyrightpermission[1]{} % Removes footnote with conference information in first column

\maketitle
\pagestyle{plain}
\thispagestyle{firstpage}

%%%%%%%%%%%%%%%%%%%%%%%%%%%%%%%%%%%%%%%%%%%%%%%%%%%%%%%%%%%
%%%%%%%%%%%%%%%%%%%%%%%%%%%%%%%%%%%%%%%%%%%%%%%%%%%%%%%%%%%
\section{Introduction}
\label{Sec_Intro}

To address the increasing demands for autonomous agents that can adapt to diverse real-world environments, simultaneous multi-task learning capabilities are highly required~\cite{mt_rl}.
The reason is that, such capabilities may enable resource-constrained autonomous agents to solve multiple tasks with limited compute resource, memory, and battery capacity. 
Moreover, real-world environments typically also demand these agents to learn from input data stream, thereby necessitating the requirements of temporal information processing capabilities~\cite{bib296, bib297, Ref_Minhas_SurveyNCL_Access55, Ref_Minhas_Replay4NCL_DAC25}.
\begin{figure}[t]
    \centering
    \includegraphics[width=0.925\linewidth]{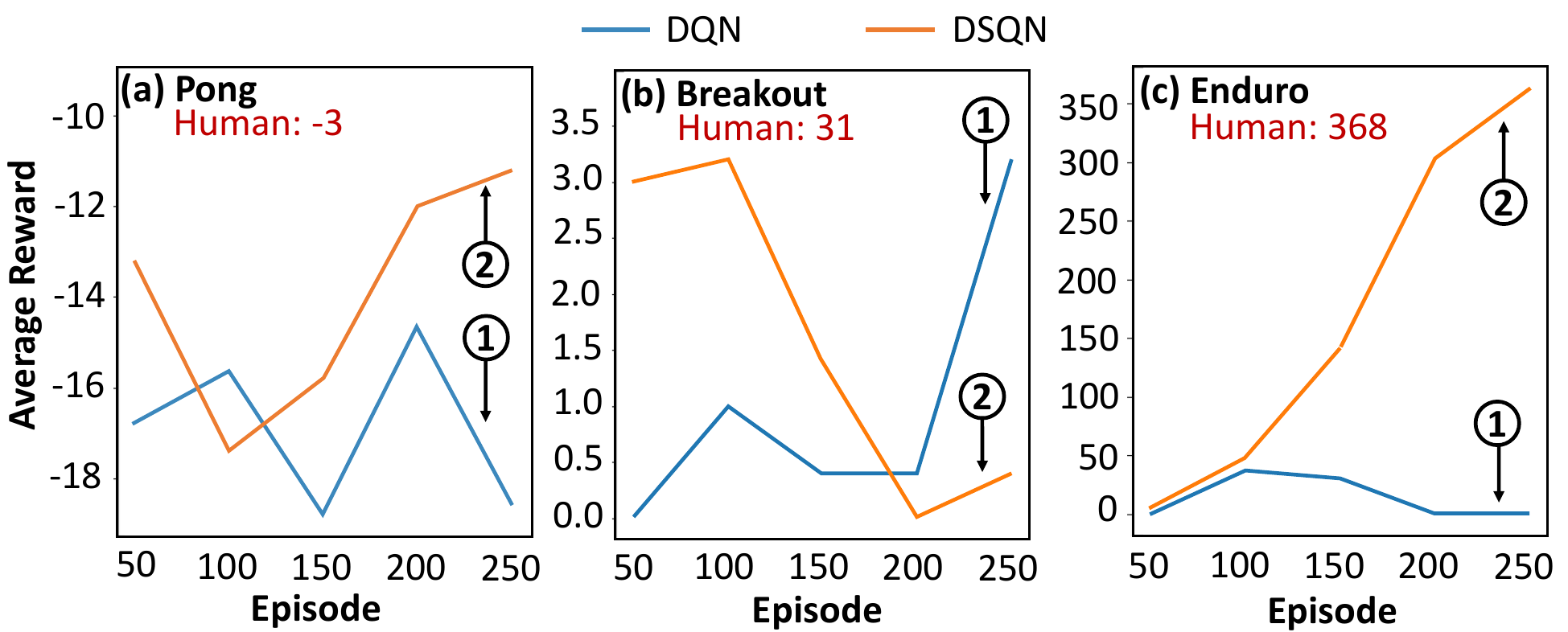}
    \vspace{-0.4cm}
    \caption{Multi-task learning performance of the state-of-the-art RL methods for ANNs (i.e., DQN~\cite{Playing_Atari}) and SNNs (i.e., DSQN~\cite{DSQN}) on three Atari games: Pong, Breakout, and Enduro. 
    }
    \label{Fig_Observation}
    \vspace{-0.6cm}
\end{figure}

In recent years, reinforcement learning (RL)-based neural network (NN) methods have shown remarkable advancements in training agents to solve complex tasks autonomously. 
For instances, such RL-based NNs are employed for mastering Go, Chess, or Atari games~\cite{Playing_Atari, Mastering_Chess_Shogi, Human_Level_Control}.
Therefore, RL-based NN methods are widely considered for solving multi-task learning challenges~\cite{MTSpark}.
Despite their advancements, the state-of-the-art RL-based NN methods still suffer from \textit{task interference}, where conflicting objectives from different tasks disturb the learning process, hence leading to sub-optimal performance in solving multiple tasks~\cite{MTSpark}.
These limitations are illustrated in Fig.~\ref{Fig_Observation}, where the state-of-the-art RL-based methods in Artificial Neural Network (ANN) and Spiking Neural Network (SNN) struggle to effectively learn multiple tasks, i.e., three different Atari games: Pong, Breakout, and Enduro. 
Specifically, the ANN-based \textit{Deep Q-Network} (DQN)~\cite{Playing_Atari} can progress relatively well in learning Breakout, while suffering performance degradation in other games; as shown by~\textcircled{\raisebox{-0.3ex}1}. 
Meanwhile, the SNN-based \textit{Deep Spiking Q-Network} (DSQN)~\cite{DSQN} can progress well in learning Pong and Enduro, while suffering from performance degradation in Breakout; as shown by~\textcircled{\raisebox{-0.3ex}2}. 
This condition limits the practicality of employing RL-based autonomous agents in real-world applications where generalization across tasks is critically needed~\cite{mt_rl}.

\textbf{Research Problem:} 
\textit{If and how can we enable multi-task learning capabilities that can progress well over time (e.g., increasing the performance over training episodes) across all given tasks?}
An efficient solution to this problem provides a scalable multi-task learning strategy and enables on-device learning capabilities, that pave the way toward intelligent autonomous agents.

%%%%%%%%%%%%%%%%%%%%%%
\vspace{-0.1cm}
\subsection{State-of-the-Art for RL-based Multi-Task Learning and Their Limitations}
\label{Sec_Intro_SOTA}

In ANN domain, RL-based multi-task learning primarily explores \textit{replay-based} and \textit{architectural-based methods}~\cite{Ref_Minhas_SurveyNCL_Access55}. 
Replay methods revisit data from previous tasks to avoid forgetting old knowledge at the cost of significant storage costs~\cite{Replay1, Replay2}, while architectural methods dynamically expand the network, leading to high latency and uncontrolled model growth~\cite{Ref_Minhas_SurveyNCL_Access55}. 
These methods show promising results, but require substantial computational resources as well as still struggle with efficiency and scalability~\cite{Ref_Minhas_SurveyNCL_Access55}. 
Another method learns from a shared initialization to quickly adapt to new tasks with minimal fine-tuning~\cite{finn2017modelagnosticmetalearningfastadaptation}, but requires an effective initialization which is a non-trivial problem.

Currently, the prominent work in ANN domain employs DQN, which performs well in a single-task setting, but still struggles in multi-task settings, as shown by \textcircled{\raisebox{-0.3ex}1} in Fig.~\ref{Fig_Observation}. 
This shows the difficulty of applying traditional RL-based models to more complex, multi-task scenarios.
In the SNN domain, several methods have been proposed.
First is DSQN~\cite{DSQN}, which provides an SNN equivalent to the DQN. 
In multi-task settings, DSQN achieves better performance than DQN due to better temporal information processing capabilities; see \textcircled{\raisebox{-0.3ex}2} of Fig.~\ref{Fig_Observation}. 
However in general, DSQN still suffers from low performance in Breakout. 
Recently, MTSpark~\cite{MTSpark} outperforms other RL-based methods from both ANN and SNN domains by achieving state-of-the-art results (i.e., Pong: -5.4, Breakout: 0.6, and Enduro: 371.2), reaching human-level performance (i.e., Pong: -3, Breakout: 31, and Enduro: 368). 
Hence, \textit{MTSpark represents the current state-of-the-art for RL-based multi-task learning methods}. 

Despite performance improvements, most of the existing RL-based methods, including the state-of-the-art MTSpark, rely on \textit{``fixed task-switching intervals''} during their training phase, thereby limiting their adaptability and efficiency in real-world dynamic environments. 
The reason is that, \textit{fixed task-switching intervals may lead to inefficient resource utilization if a task has already plateaued during training or if a task requires more time of training}.      
Moreover, most of multi-task RL methods rely on separate task training and testing. It is different from the targeted problem in this work, i.e., \textit{simultaneous multi-task RL which trains different tasks at the training phase, thus developing only one model for solving multiple tasks}.

%%%%%%%%%%%%%%%%%%%%%%%%%%%%%%%%%%%%%%%%%
\vspace{-0.2cm}
\subsection{Associated Research Challenges}
\label{Sec_Intro_Challenges}

To solve the limitations posed by the fixed task-switching intervals and improve the effectiveness of multi-task learning, the following \textbf{key research challenges} need to be addressed.
\begin{itemize}[leftmargin=*]
    \item The network model should include an adaptive mechanism to monitor task-specific learning progress and decide when to switch tasks. 
    This will help the network to have scalable multi-task learning capabilities.
    \item The task-switching mechanism should avoid both premature and delayed task transitions during training, hence ensuring each task receives adequate training time.
    \item The network model should maintain its complexity level (e.g., network size) when performing the algorithms for multi-task learning, thereby ensuring that the network can be efficiently run in resource-constrained autonomous agents.
\end{itemize}

%%%%%%%%%%%
\vspace{-0.2cm}
\subsection{Our Novel Contributions}
\label{Sec_Intro_Novelty}

\begin{figure}[h]
    \vspace{-0.3cm}
    \centering
    \includegraphics[width=0.85\linewidth]{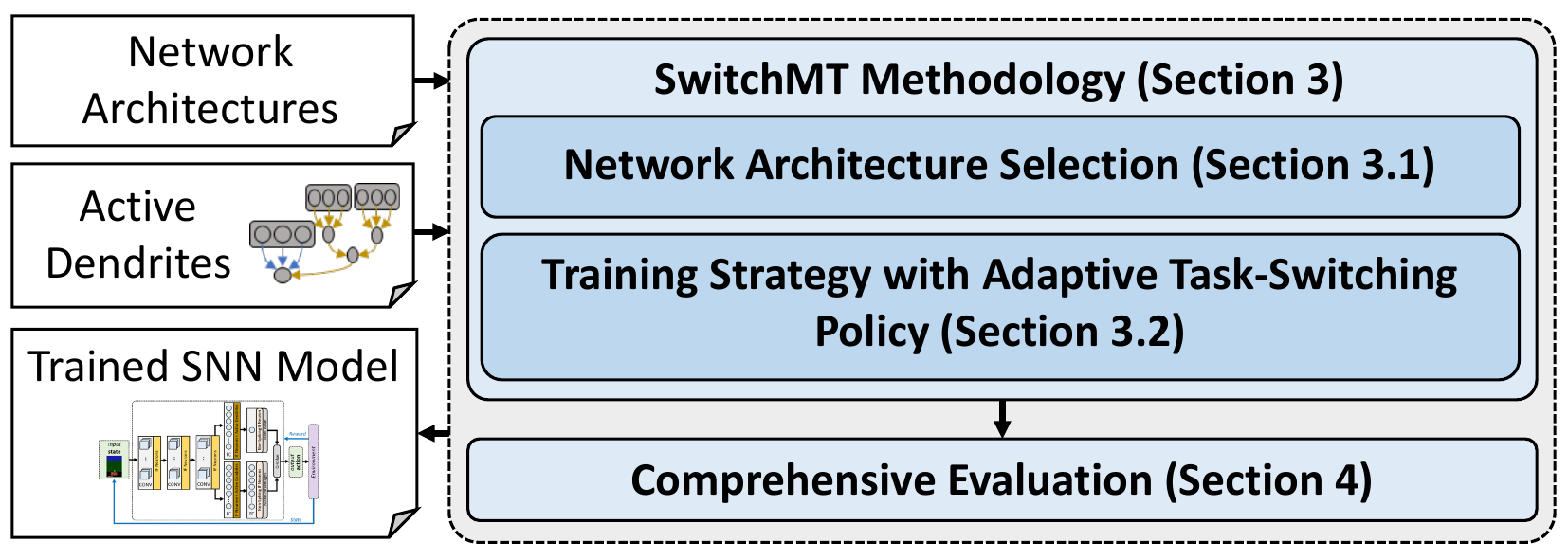}
    \vspace{-0.4cm}
    \caption{Overview of our novel contributions in this work.}
    \label{Fig_Novelty}
    \vspace{-0.6cm}
\end{figure}

To address the research challenges, we propose \textbf{SwitchMT}, \textit{a novel methodology to enable scalable and simultaneous multi-task learning through an adaptive task-switching policy in SNN-based solutions for intelligent autonomous agents}; see an overview in Fig.~\ref{Fig_Novelty}. 
To achieve this, we employ following key contributions.
\begin{itemize}[leftmargin=*]
    \item \textbf{Network architecture selection (Section~\ref{Sec_SwitchMT_Network}):} 
    It chooses the suitable network architecture that effectively facilitates RL-based multi-task learning.
    Here, we leverage a deep spiking Q-network (DSQN) architecture with active dendrites and dueling structure.
    \item \textbf{Training strategy with an adaptive task-switching policy (Section~\ref{Sec_SwitchMT_Training}):} 
    It targets to automatically adjust task-switching decisions based on real-time performance monitoring by leveraging both rewards and internal dynamics of network parameters, thus leading to more effective training.
    \item \textbf{Comprehensive evaluation (Section~\ref{Sec_Eval}):} 
    We perform extensive evaluation including three Atari games (i.e., Pong, Breakout, and Enduro), and compare the performance with several prominent and state-of-the-art works (i.e., DQN, DSQN, and MTSpark\_ADD) in terms of both rewards and game points. 
\end{itemize}
\textbf{Key Results:}
We evaluate the SwitchMT methodology using a Python-based implementation, and then run it on the Nvidia GeForce RTX 4090 multi-GPU machines. 
It achieves competitive performance scores across multiple Atari games (i.e., Pong: -8.8, Breakout: 5.6, and Enduro: 355.2) as compared to the state-of-the-art MTSpark (i.e., Pong: -5.4, Breakout: 0.6, and Enduro: 371.2).
Moreover, SwitchMT also obtains higher game points and longer game episodes than the state-of-the-art without increasing network complexity, thus reflecting its effectiveness for multi-task learning. 

%%%
\begin{figure*}[t]
    \centering
    \includegraphics[width=0.85\linewidth]{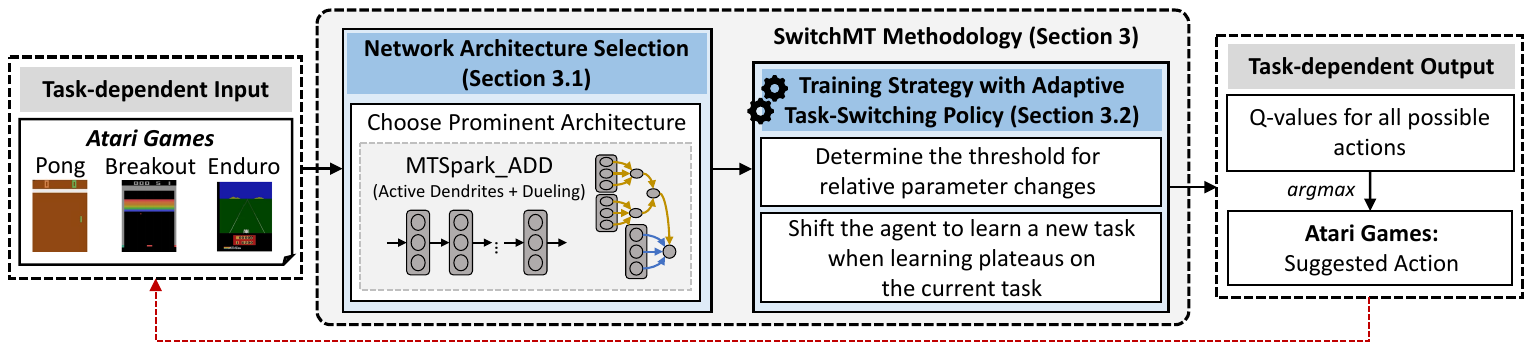}
    \vspace{-0.4cm}
    \caption{Our SwitchMT methodology and its key steps: network architecture selection and adaptive task-switching policy.}
    \label{Fig_SwitchMT}
    \vspace{-0.4cm}
\end{figure*}
%%%

%%%%%%%%%%%%%%%%%%%%%%%%%%%%%%%%%%%%%%%%%%%%%%%%%%%%%%%%%%%
%%%%%%%%%%%%%%%%%%%%%%%%%%%%%%%%%%%%%%%%%%%%%%%%%%%%%%%%%%%
\section{Background}
\label{Sec_Back}

\subsection{Spiking Neural Networks (SNNs)}
\label{Sec_Back_SNNs}

Biologically-inspired SNNs have emerged as an energy-efficient alternative for machine learning (ML) algorithm, due to their event-based operations, which can lead to energy-efficient and temporally precise computation~\cite{MAASS19971659, Pfeiffer2018DeepLW}.
An SNN model consists of several design components, including network architecture, spiking neurons, neural coding, and learning rule~\cite{Ref_Mozafari_Spyketorch_FNINS19}.
In recent years, SNNs have demonstrated their potentials for achieving high performance in solving many ML-based applications, such as image classification~\cite{Active_Image}, automotive~\cite{Ref_Cordone_ObjDetSNN_IJCNN22}, healthcare~\cite{Ref_Luo_EEGSNN_Access20}, and robotics~\cite{Ref_Bartolozzi_EmbodiedNeuroIntel_Nature22}. 
These SNN models may be developed through manual design~\cite{Ref_Diehl_SNN_FNCOM15} or neural architecture search~\cite{Ref_Putra_SpikeNAS_TAI25}. 
SNN ability to efficiently process temporal information is suitable for processing data stream, which is an important feature for multi-task learning.
For examples, recent studies have leveraged temporal information within spike trains and bio-plausible learning rules to improve performance in sequential task learning~\cite{bib296, bib297, bib72}, but they mostly focused on relatively simple tasks, such as the MNIST dataset.
Meanwhile, in the RL realm, DSQN~\cite{DSQN} and MTSpark~\cite{MTSpark} have been recently proposed to facilitate the learning process through spiking Q-network architecture, while capturing temporal dynamics and their potential for low-power implementation.  

%%%%%%%%%%%%%%%%%%%%%%%%%%%%%%%%%%%%%%%%%
\vspace{-0.2cm}
\subsection{Methods for Mitigating Task Interference}
\label{Sec_Back_TaskInference}

\subsubsection{Overview}
The main challenge in multi-task learning is task interference, where the acquisition of new skills degrades performance on previously learned tasks due to conflicting objectives. 
Toward this, several strategies have been developed to address this challenge, and they can be categorized into three groups: architectural-based, replay-based, and regularization-based methods~\cite{MTSpark}.
\textit{Architectural-based methods} often incorporate task-specific components such as additional layers or multi-head structures to isolate and preserve task-relevant features~\cite{Archi2, PNN}.
\textit{Replay-based methods} combat interference by interleaving new task training with revisiting samples from earlier tasks, leading to increased computational and memory demands~\cite{Replay1, Replay2, GenReplay}.
\textit{Regularization-based methods}, like Elastic Weight Consolidation (EWC)~\cite{EWC}, constrain parameter updates by penalizing changes in weights that are critical for earlier tasks.
Despite offering performance improvements, these methods introduce extra hyper-parameters and additional model complexity, which complicate the training process and can hinder scalability in large multi-task environments.

\vspace{-0.1cm}
\subsubsection{MTSpark Architecture and Its Limitations}
\label{Sec_Back_MTwithSNN}

MTSpark~\cite{MTSpark} represents the most recent advances of RL-based multi-task learning with SNNs, leveraging their temporal processing and efficient event-based operations. 
MTSpark enhances the DSQN architecture~\cite{DSQN} with two design variants: \textit{MTSpark\_AD} and \textit{MTSpark\_ADD}. 
Specifically, MTSpark\_AD enhances the DSQN architecture with active dendrites, while MTSpark\_ADD enhances with both active dendrites and dueling structure. 
Here, \textit{active dendrites} employ task-specific context signals to dynamically modulate neuronal activations to create specialized sub-networks within a network model.
Meanwhile, \textit{dueling structure} (with separate state value and action advantage estimators) improves the ability to generalize learning across actions without algorithm modification~\cite{MTSpark}.
In general, MTSpark\_ADD performs better than MTSpark\_AD due to better task-based sub-network specialization. 
A critical drawback in MTSpark is its \textit{fixed task-switching interval}, i.e., training for 25 episodes in each environment (task) before switching. 
Moreover, this rigid schedule does not account for the learning progress of the agent. 
In some environments, we observe that the agent may reach learning plateau much earlier, leading to wasted training episodes with minimal returns. 
Meanwhile in more challenging settings, the selected fixed interval may be insufficient for providing adequate learning. 
Consequently, \textit{fixed task-switching interval can lead to inefficient training or contribute to overfitting within specific tasks}.

%%%%%%%%%%%%%%%%%%%%%%%%
%%%%%%%%%%%%%%%%%%%%%%%%
\vspace{-0.2cm}
\section{The SwitchMT Methodology}
\label{Sec_SwitchMT}

The key steps of our SwitchMT methodology are illustrated in Fig.~\ref{Fig_SwitchMT}, which are described in the following sub-sections.

%%%%%%%%%%%%%%%%%%%%%%%%%%%%%%%%%%%%%%%%%%%%%%%%
\vspace{-0.2cm}
\subsection{Network Architecture Selection}
\label{Sec_SwitchMT_Network}

This step aims to select the suitable network architecture that can facilitate RL-based multi-task learning. 
Here, we select the MTSpark\_ADD architecture~\cite{MTSpark}, which is based on DSQN with active dendrites and dueling structure, since it has demonstrated the state-of-the-art performance, as discussed in Section~\ref{Sec_Back_MTwithSNN}. 
Details of this architecture are discussed below. 

%%%%%%%%%%%%%%%%%
\vspace{-0.2cm}
\subsubsection{Spiking Neuron Model with Active Dendrites}

\begin{figure}[h]
    \vspace{-0.2cm}
    \centering
    \includegraphics[width=0.65\linewidth]{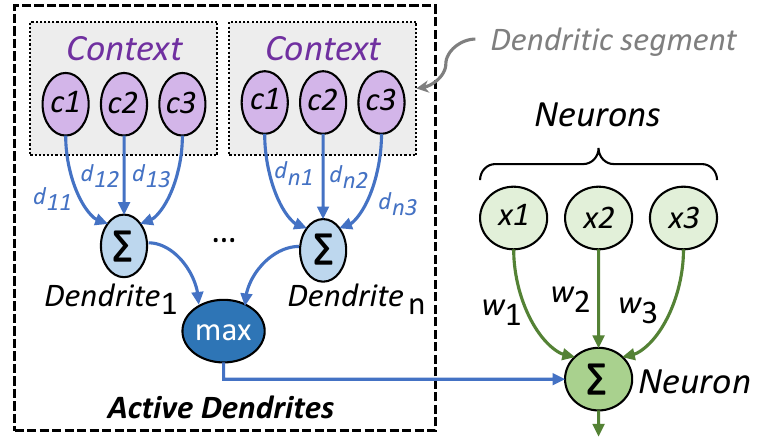}
    \vspace{-0.4cm}
    \caption{The integrate-and-fire (IF) neuron model is enhanced using active dendrites with context signals; adapted from MTSpark\_ADD architecture~\cite{MTSpark}.}
    \label{Fig_ActiveDendrites}
    \vspace{-0.6cm}
\end{figure}

We enhance multi-task learning by employing \textit{adaptive dendrites}, which dynamically modulate integrate-and-fire (IF) neurons based on task context, as illustrated in Fig.~\ref{Fig_ActiveDendrites}. 
This allows neurons to selectively respond to different tasks, reducing task interference and improving efficiency.
Here, the neuron's membrane potential update incorporates a context-sensitive dendritic modulation, as formulated in Eq.~\ref{eq_NeuronWithDendrites}.
\vspace{-0.1cm}
\begin{equation}
\vspace{-0.1cm}
    V(t) = V(t - \Delta t) + f\left(\sum_{i}s_{i}(t), \max_{j}(d_j^Tc)\right)
    \label{eq_NeuronWithDendrites}
\end{equation}
\noindent where $d_j$ are dendritic weights, $s_i(t)$ represents presynaptic spikes, and $c$ is the context signal. 
A neuron spikes when $V(t)$ exceeds the threshold potential ($V_{th}$). 
This forms task-adaptive sub-networks by dynamically strengthening or suppressing specific pathways based on task demands.

%%%%%%%%%%%%%%%%%
\subsubsection{Network Architecture}

\begin{figure}[t]
    \centering
    \includegraphics[width=0.9\linewidth]{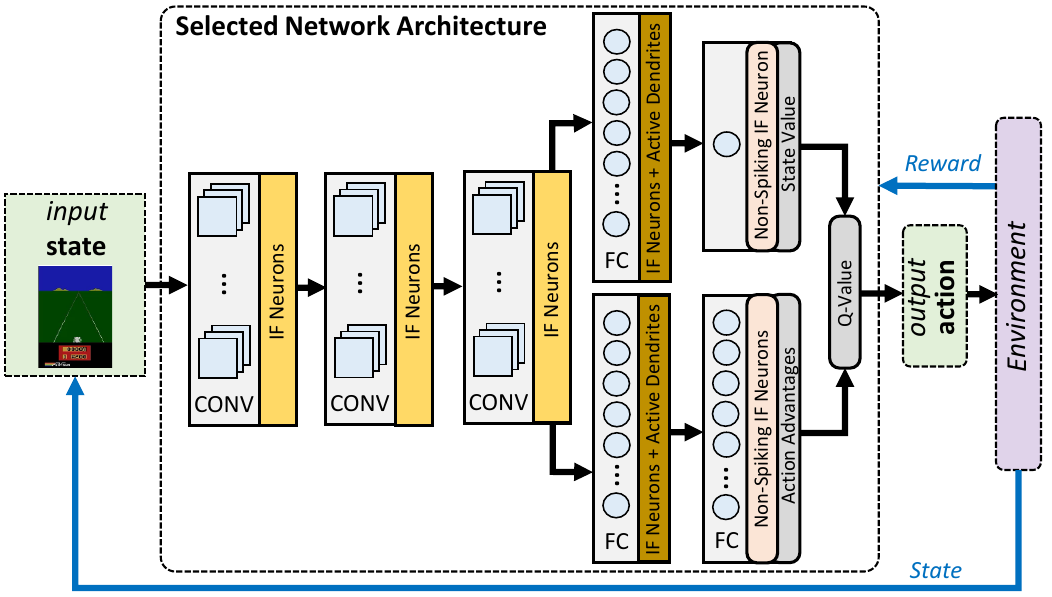}
    \vspace{-0.4cm}
    \caption{The network architecture used in our proposed SwitchMT methodology, employs active dendrites and dueling structure; based on the MTSpark\_ADD architecture~\cite{MTSpark}.}
    \label{Fig_SwitchMT_Arch}
\end{figure}

The configuration of selected network architecture is illustrated in Fig.~\ref{Fig_SwitchMT_Arch}. 
It consists of three convolutional (CONV) layers with kernel sizes of 8, 4, and 3 and strides of 4, 2, and 1, respectively. 
Each CONV layer is followed by batch normalization (BatchNorm) and an IF neuron layer. 
Input data streams are processed through these layers, generating feature maps that are flattened and then fed to a fully connected (FC) layer with 512 units. 
Afterward, the output is fed into a spiking neuron layer equipped with adaptive dendrites and context signals, which modulates activity based on task context. 
A final FC layer produces Q-values over all possible actions.
Furthermore, the dueling structure employs two estimators: \textit{a state-value function} that estimates the overall value of a given state, and \textit{an action-specific advantage function} that determines the relative importance of each action in that state. 
This function separation improves generalization across actions, allowing the model to better differentiate the impact of individual actions in varying states.

%%%%%%%%%%%%%%%%%%%%%%%%%%%%%%%%%%%%%%%%%%%%%%%%
\vspace{-0.2cm}
\subsection{Training Strategy with Our Novel Adaptive Task-Switching Policy}
\label{Sec_SwitchMT_Training}

%%%%%%%%%%%%%%%%%
\subsubsection{Training Strategy}

This step aims at devising an adaptive task-switching policy and integrate it into a consolidated training strategy. 
To do this, we propose the training algorithm presented in Algorithm~\ref{alg:adaptive_alg}, while considering the following design choices.

\begin{itemize}[leftmargin=*]
    \item We maintain a fixed size replay buffer ($N$) to efficiently store transitions for each environment.
    \item We linearly anneal the exploration rate $\epsilon$ from an initially high rate to a low rate over the first $T$ frames, and keep it fixed at the low rate, hence allowing the agent to freely explore possible actions in earlier frames and gradually steer the agent toward more effective actions later.
    \item We define our target ($y_n$) as expressed in Equation~\ref{eq_NetworkTarget}, where $r_n$ is the reward received after taking an action in the state $\phi_n$, $\phi_{n + 1}$ is the state seen immediately after taking the action, $Q$ is our network, and $Q^*$ is the target network.
    \begin{equation}
    y_n = \begin{cases}
        r_n; \text{~if $\phi_{n+1}$ is terminal}\\
        r_n + \gamma Q^{*}(\phi_{n + 1}, \arg\max_{a} Q(\phi_{n + 1}, a)); \text{~otherwise}
    \end{cases}
    \label{eq_NetworkTarget}
    \end{equation}
    \vspace{-0.3cm}
\end{itemize}

%%%%%%%%%%%%%%%%%
\vspace{-0.3cm}
\subsubsection{Adaptive Task-Switching Policy}

To address the limitations of fixed-interval environment switching, we introduce a dynamic strategy that automatically shifts the agent to a new task when learning plateaus on the current task, called \textit{\textbf{adaptive task-switching policy}}. 
This is determined by monitoring the relative change in model parameters over a sliding window of 
$K$ episodes. 
If the change falls below a predefined threshold (e.g., 10\%), the agent switches environments, ensuring continued learning progress.
Formally, this policy can be expressed as Equation~\ref{eq_DeltaTheta}.
Here, $\theta_t$ denotes the model parameters at the end of episode $t$. 
After each episode, we compute the relative parameter change ($\Delta\theta$) over the last $K$ episodes using the L2 norm technique.
\begin{equation}
    \Delta \theta = \frac{||\theta_t - \theta_{t-K}||_2}{||\theta_{t-K}||_2} \times 100%.
    \label{eq_DeltaTheta}
\end{equation}

\begin{algorithm}[t]
    \caption{Our proposed training strategy}
    \label{alg:adaptive_alg}
    \begin{algorithmic}[1]
    \footnotesize
    \STATE \textbf{Initialization:}
    \STATE \;\;\; Network $Q$, target network $Q^*$
    \STATE \;\;\; Environments $E$, replay buffers $D_j$
    \STATE \;\;\; Environment index $i \gets 0$, parameter history buffer $H_i \gets []$
    \STATE \textbf{Process:}
    \FOR{timestep $t = 1$ to $T_{\text{max}}$}
        \STATE Select action $a$ via $\epsilon$-greedy policy
        \STATE Execute action $a$, observe reward $r$ and next state $\phi'$
        \STATE Store transition $(\phi, a, r, \phi')$ in $D_{i}$
        \STATE Sample transitions $(\phi_n, a_n, r_n, \phi_{n+1})$ from $D_{i}$
        \STATE Set target $y_n$ as defined in Equation \ref{eq_NetworkTarget}
        \STATE Perform gradient descent on $(y_n - Q(\phi_n, a_n))^{2}$
        \IF{\(\phi'\) is terminal}
            \STATE Reset \(\phi\), append \(\theta_t\) to \(H_i\)
             \IF{\(|H_i| > K\)}
                \STATE Remove oldest entry from \(H_i\)
            \ENDIF
            \IF{\(|H_i| = K\)}
                \STATE Compute \(\Delta \theta\) using Equation~\ref{eq_DeltaTheta}
                \IF{\(\Delta \theta < 10\%\)}
                    \STATE \(i \gets (i + 1) \mod |E|\), \(H_i \gets []\)
                \ENDIF
            \ENDIF
            \ELSE
                \STATE Update \(\phi \gets \phi'\)
        \ENDIF
        \IF{\(t \mod \text{target\_update} = 0\)}
            \STATE Update \(Q^* \gets Q\)
        \ENDIF
    \ENDFOR
\end{algorithmic}
\end{algorithm}
\setlength{\textfloatsep}{2pt}

The proposed adaptive task-switching policy has the following characteristics and advantages.
\begin{itemize}[leftmargin=*]
    \vspace{-0.1cm}
    \item \textit{\textbf{Task-Agnostic Adaptation}}: 
    Fixed task-switching interval in prior work makes two implicit assumptions: (1) all tasks require equal training time, and (2) learning progress is linear across episodes. 
    However, we observe that neither of those assumptions hold in practice. 
    Complex tasks (e.g., sparse-reward environments) often need extended training, while simple tasks can eliminate the need of extended training. 
    Through the proposed adaptive task-switching policy, our adaptive task-witching policy automatically detects when parameter updates stagnate (i.e., $\Delta\theta < 10\%$), hence spending more time on harder tasks without manual tuning.
    \item \textit{\textbf{Cross-Task Generalization}}: 
    Fixed interval may force premature task-switching on unfinished tasks, causing \textit{catastrophic interference}, since insufficient learning leaves task-specific patterns under-encoded in dendritic weights. It may also cause \textit{overfitting} due to extended training on mastered tasks wastes capacity on redundant updates. 
    In contrast, our adaptive task-witching policy avoids both extremes by tying the switching decision to the parameter changes, i.e., tasks are retained until their marginal learning returns diminish.
\end{itemize}

%%%%%%%%%%%%%%%%%%%%%%%%
%%%%%%%%%%%%%%%%%%%%%%%%
\vspace{-0.3cm}
\section{Evaluation Methodology}
\label{Sec_Eval}

\begin{figure}[h]
    \vspace{-0.1cm}
    \centering
    \includegraphics[width=0.95\linewidth]{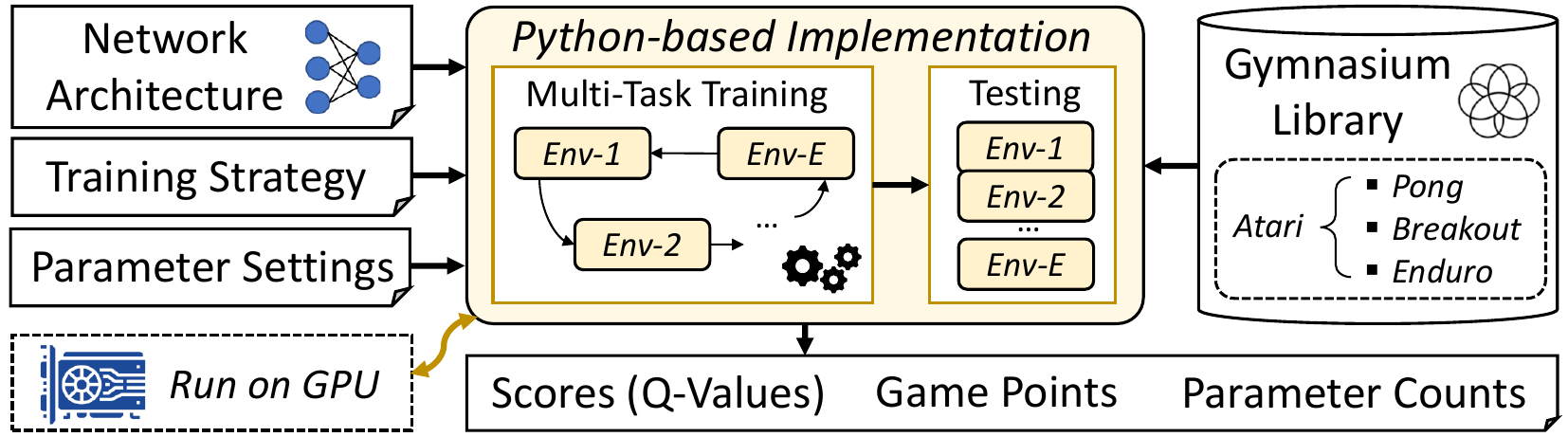}
    \vspace{-0.4cm}
    \caption{Experimental setup used in the evaluation.}
    \label{Fig_ExpSetup}
    \vspace{-0.3cm}
\end{figure}

We evaluate the SwitchMT methodology through Python implementation based on the Gymnasium library~\cite{Ref_Towers_Gym_arXiv24}, then run it on the NVIDIA GeForce RTX 4090 multi-GPU machines, as shown in Fig.~\ref{Fig_ExpSetup}.
For comparison partners, we consider DQN~\cite{Playing_Atari}, DSQN~\cite{DSQN}, DQN with dueling structure (i.e., DQN\_D)~\cite{Dueling_DQN}, DSQN with dueling structure (i.e., DSQN\_D)~\cite{MTSpark}, and MTSpark\_ADD~\cite{MTSpark}.
DQN and DSQN represent the baseline RL-based methods, DQN\_D and DSQN\_D represent the structurally-enhanced methods, while MTSpark\_ADD represents the state-of-the-art.
All these methods consider a fixed task-switching interval (i.e., 25 episodes in each environment).
An episode means a sequence of frames that starts from the initial state of the game and ends when the terminal condition is met, such as when the player ``loses all lives'' or the game ends.
Outputs of the experiments include the performance scores (i.e., Q-values), game points, and number of trainable parameters.
Scores (Q-values) represent how good the given method generates next actions, game points represent how the actual game progress visually, and number of trainable parameters represents network size.

% \smallskip
\textbf{Targeted Tasks (Environments):}
We train and test all investigated methods on three Atari games (i.e., Pong, Breakout, and Enduro) as the target environments, and utilize an action space of 18 possible actions. 
This approach ensures that we have consistency in the action space across all environments. 
The state observations are represented as 210$\times$160$\times$3 RGB images, identical to the frames presented to human players.
We select these three Atari games to ensure a more controlled evaluation of our method, especially for observing task interference and training dynamics, while providing diversity in task's difficulty and characteristics.

% \smallskip
\textbf{Experimental Settings:}
We train all networks with a static learning rate of 1e-4, a batch size of 512, and the Adam optimizer over 4 million frames in each environment. 
In line with the widely-used settings in Deep RL research, we apply a discount factor $\gamma$ of 0.99 to stabilize learning, use a replay buffer of size $2^{20}$ for each environment to store transitions for experience replay, and decay exploration from $\epsilon$ = 1.00 to $\epsilon$ = 0.10 over 1 million total frames. The details of the parameter settings are provided in Table \ref{table:exp_setup}.

\begin{table}[h]
\vspace{-0.2cm}
\caption{Parameter settings for the experimental setup.}
\vspace{-0.3cm}
\small
\centering
\label{table:exp_setup}
\begin{tabular}{@{}cc@{}}
\toprule
\textbf{Variable}     & \textbf{Value Used} \\ 
\midrule
Replay Buffer Size    & $2^{20}$            \\
Batch Size            & 512                 \\
Epsilon Start         & 1.00                \\
Epsilon End           & 0.10                \\
Decay Epsilon Over    & 1,000,000 frames    \\
Update Target Network & Every 10,000 frames \\
Learning Rate         & 1e-4                \\
\bottomrule
\end{tabular}
\end{table}
\setlength{\textfloatsep}{2pt}

%%%%%%%%%%%%%%%%%%%%%%%%
%%%%%%%%%%%%%%%%%%%%%%%%
\vspace{-0.3cm}
\section{Results and Discussion}
\label{Sec_Results}

\begin{figure}[t]
    \centering
    \includegraphics[width=\linewidth]{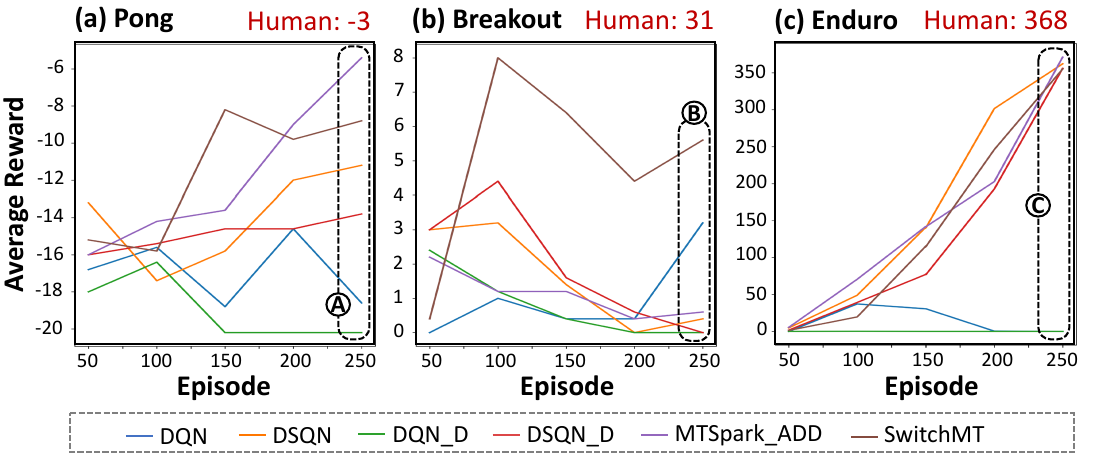}
    \vspace{-0.7cm}
    \caption{Performance of different models (i.e., DQN~\cite{Playing_Atari}, DSQN~\cite{DSQN}, DQN\_D~\cite{Dueling_DQN}, DSQN\_D~\cite{MTSpark}, MTSpark\_ADD~\cite{MTSpark}, and our SwitchMT) across Pong, Breakout, and Enduro.}
    \label{fig:perf_results}
    \vspace{-0.2cm}
\end{figure}

\begin{figure*}[t]
    \centering
    \includegraphics[width=\linewidth]{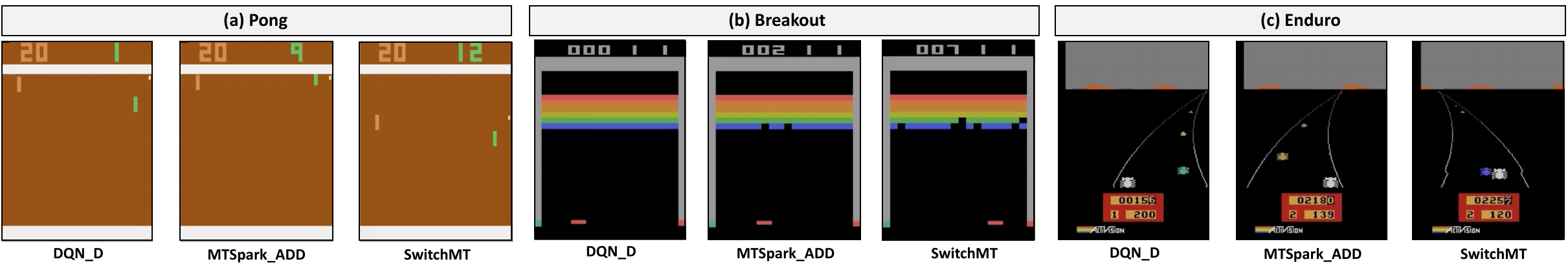}
    \vspace{-0.7cm}
    \caption{Performance evaluation of DQN\_D~\cite{Dueling_DQN}, MTSpark\_ADD~\cite{MTSpark}, and SwitchMT on \textbf{(a)} Pong, \textbf{(b)} Breakout, and \textbf{(c)} Enduro. Here, a higher game point means better performance, and often means longer game episodes.}
    \label{fig_PBE}
    \vspace{-0.2cm}
\end{figure*}

\begin{table*}[t]
\caption{Performance of network models from different methods across different tasks (i.e., Pong, Breakout, and Enduro). 
SwitchMT achieves high scores for all tasks, while previous works fail to do so; showing the superior performance of SwitchMT.}
\vspace{-0.3cm}
\small
\centering
\begin{tabular}{c|cccccc|c}
\toprule
\textbf{Environment} & \textbf{DQN~\cite{Playing_Atari}} & \textbf{DSQN~\cite{DSQN}} & \textbf{DQN\_D~\cite{Dueling_DQN}} & \textbf{DSQN\_D~\cite{MTSpark}} & \textbf{MTSpark\_ADD~\cite{MTSpark}} & \textit{\textbf{SwitchMT}} & \textbf{Human~\cite{Playing_Atari}} \\ 
\midrule
\midrule
Pong & -18.6 & -11.2 & -20.2 & -13.8 & -5.4 & -8.8 & -3 \\ 
Breakout & 3.2 & 0.4 & 0 & 0 & 0.6 & 5.6 & 31 \\
Enduro & 0 & 362.2 & 0 & 356 & 371.2 & 355.2 & 368 \\ \bottomrule
\end{tabular}
\label{perf_table}
\vspace{-0.3cm}
\end{table*}
\setlength{\textfloatsep}{6pt}

%%%%%%%%%%%%%%%%%%%%%%%%%%%%%%%%%%%%%%%%%
\subsection{Performance on Atari Games}

Experimental results for performance evaluation are provided in Fig.~\ref{fig:perf_results} and Table~\ref{perf_table}.  
These results show that, performance of SwitchMT is competitive with the state-of-the-art MTSpark\_ADD and outperforms the conventional RL-based methods without active dendrites (i.e., DQN, DSQN, DQN\_D, and DSQN\_D). 
These show the positive impact of active dendrites, dueling structure, and our proposed adaptive task-switching policy on the performance.
Specifically, \textit{active dendrites and dueling structure help in identifying the context and task-specific sub-networks during the training phase}.
Meanwhile, \textit{our adaptive task-switching policy helps in identifying the effective time of switching the training environment}, avoiding premature transitions from unfinished tasks, and unnecessary extended training or delayed transitions from mastered tasks, ensuring adequate training for each task.
We also make several key observations below. 
\begin{itemize}[leftmargin=*]
    \item \textbf{Pong}: 
    SwitchMT achieves a score of -8.8, surpassing models from DQN (score: -18.6) and DSQN (score: -11.2), while approaching human-level performance (score: -3) and remaining competitive with the MTSpark\_ADD (score: -5.4); see label~\textcircled{\raisebox{-0.3ex}A}. 
    In contrast, models from other methods struggle to learn effectively, with scores ranging from -20.2 to -11.2. 
    Fig.~\ref{fig_PBE}(a) illustrates the final game points obtained by the trained DQN\_D, MTSpark\_ADD, and SwitchMT on Pong. 
    Among these results, SwitchMT achieves the most competitive point (with 12 points), outperforming both DQN\_D (with 1 point) and MTSpark\_ADD (with 9 points). 
    Here, active dendrites enable the model in SwitchMT to perform well across environments, while adaptive task-switching policy further enhances performance due to timely switching decision. 
    \item \textbf{Breakout}: 
    SwitchMT achieves a score of 5.6, outperforming all other methods such as DQN (score: 3.2), DSQN (score: 0.4), and MTSpark (score: 0.6), but it is still below human-level performance (score: 31); see~\textcircled{\raisebox{-0.3ex}B}. 
    These results highlight the challenges faced by all network models in this environment. 
    Therefore, Breakout remains a difficult task where no model achieves human-level performance, similar to findings in previous studies~\cite{MTSpark}. 
    While all other methods collapse (e.g., DQN\_D achieves 0 point and MTSpark\_ADD achieves 2 points), our SwitchMT attains the highest performance (with 7 points), as shown in Fig.~\ref{fig_PBE}(b). 
    SwitchMT exhibits an emergent strategy of directing the ball toward the edges, similar to behaviors observed in~\cite{Playing_Atari}. 
    \item \textbf{Enduro}: 
    SwitchMT achieves a score of 355.2, closely matching human performance (score: 368), DSQN (score: 362.2), and MTSpark (score: 371.2), while outperforming models like DQN (score: 0); see~\textcircled{\raisebox{-0.3ex}C}.  
    These results highlight the effectiveness of both SwitchMT and MTSpark in adapting to the Enduro environment. 
    Fig.~\ref{fig_PBE}(c) illustrates the final game points obtained by trained DQN\_D, MTSpark\_ADD, and SwitchMT on Enduro. 
    Among these results, SwitchMT and MTSpark\_ADD achieve impressive game points, while DQN\_D once again fails to learn, since it ends on Day-1.
    For instance, at one point in time, DQN\_D, MTSpark\_ADD, and SwitchMT achieve game points of 156, 2180, and 2250, respectively.
    These results underscore the significant benefits of active dendrites, enabling effective task modulation within the networks.
    Moreover, SwitchMT maintains a slight edge over MTSpark\_ADD, achieving a slightly higher game point, highlighting the benefits of our adaptive task-switching policy. 
    \vspace{-0.2cm}
\end{itemize}
    
%%%%%%%%%%%%%%%%%%%%%%%%%%%%%%%%%%%%%%%%%
\subsection{Comparison of Network Model Sizes}

As shown in Table~\ref{table_arch_params}, parameter counts for the DQN and DSQN models are the same since they employ similar network architecture. 
Due to dueling structure, parameter counts for the DQN\_D and DSQN\_D models are about 2x from parameter counts for the DQN and DSQN models, respectively.
Meanwhile, parameter count for the MTSpark\_ADD model is nearly identical to the DSQN\_D model, and such a slight difference comes from the active dendrites. 
This shows that integrating active dendrites into IF neurons has negligible parameter overhead. 
Furthermore, parameter counts for the MTSpark\_ADD and SwitchMT models are the same, as they employ the same network architecture.
This shows that, the enhanced performance of our SwitchMT over the state-of-the-art MTSpark\_ADD comes from the proposed adaptive task-switching policy that effectively enables task-agnostic adaptation and cross-task generalization. 
\textit{These findings underscore the ability of our SwitchMT to manage multiple tasks efficiently without significantly enlarging the model, hence enabling scalable multi-task learning}.

\begin{table}[t]
  \caption{Number of trainable parameters of different network models, which reflects the corresponding model size.}
  \label{table_arch_params}
  \vspace{-0.3cm}
  \small
  \centering
  \begin{tabular}{@{}lc@{}}
    \toprule
    \textbf{Model}  & \textbf{\# Trainable Parameters} \\ \midrule
    DQN~\cite{Playing_Atari}    & 1,693,682  \\
    DSQN~\cite{DSQN}            & 1,693,682  \\
    DQN\_D~\cite{Dueling_DQN}   & 3,300,339  \\
    DSQN\_D~\cite{MTSpark}      & 3,300,339  \\
    MTSpark\_ADD~\cite{MTSpark} & 3,300,357  \\ 
    SwitchMT                    & 3,300,357  \\\bottomrule
  \end{tabular}
\end{table}
\setlength{\textfloatsep}{6pt}

%%%%%%%%%%%%%%%%%%%%%%%%%%%%%%%%%%%%%%%%%%%%%
\subsection{Ablation Study and Further Discussion}

\textbf{Impact of Active Dendrites:}
Employing active dendrites helps in providing fast accurate response and better control for long-term training with delayed rewards.
For Pong, adding active dendrites on DSQN improves the performance from -11.2 to -9.4, while adding them on DSQN\_D improves the performance from -13.8 to -5.4.
For Breakout, adding active dendrites on DSQN improves the performance from 0.4 to 2, while adding them on DSQN\_D improves the performance from 0 to 0.6.
For Enduro, adding active dendrites on DSQN improves the performance from 362.2 to 363.2, while them on DSQN\_D improves the performance from 356 to 371.2.

\smallskip
\textbf{Impact of Dueling Structure:}
Overall, dueling structure increases the model size by about $2\times$ than the non-dueling one.
In terms of performance for Pong, adding dueling structure on DQN and DSQN slightly degrades the performance from -18.6 to 20.2 as well as from -11.2 to -13.8, respectively.
However, when active dendrites are incorporated, the model with dueling structure performs better (-5.4) than the non-dueling one (-9.4).
This indicates that benefits of dueling structure may appear when coupled with active dendrites.
For Enduro, adding dueling structure on DSQN does not change the performance much, but improves the performance for MTSpark\_ADD (371.2) from 363.2.
This suggests that the dueling structure may obtain some gains when used together with active dendrite and spiking neurons.
For Breakout, adding dueling structure on DQN and DSQN slightly degrades the performance from -18.6 to 20.2 as well as from 0.4 to 0, respectively.
This suggests that the dueling structure still struggles to solve Breakout. 

\smallskip
\textbf{Reduction in the Training Time and Overfitting:}
Dynamically adjusting the transition between environments based on the agent's learning progress, makes SwitchMT moves away from fixed training schedules (e.g., 25 episodes per environment). 
Instead, it monitors the stabilization of model parameters, indicating when additional training yields diminishing returns. 
It shortens training duration and mitigates overfitting by preventing excessive training in environments where the model has achieved high performance.

\smallskip
\textbf{Dynamic Curriculum for Agents:}
SwitchMT inherently establishes a dynamic curriculum. 
Specifically, environments that continue to provide substantial learning opportunities keep the agent engaged longer. 
Simultaneously, the agent can move on from environments where learning has plateaued, ensuring efficient use of training resources.
In this manner, SwitchMT enhances efficiency and promotes better generalization across tasks.

\smallskip
\textbf{Eliminating the Task-Switching Hyperparameter Tuning:}
Prior methods often rely on a pre-determined switch interval hyperparameter~\cite{MTSpark}, which requires extensive tuning due to its sensitivity. 
Our SwitchMT eliminates this need, simplifying the training process and reducing the complexity associated with hyperparameter tuning. 
This simplification is particularly beneficial, as hyperparameter tuning can be labor-intensive and computationally expensive~\cite{eimer2023hyperparametersreinforcementlearningtune}.

%%%%%%%%%%%%%%%%%%%%%%%%%%%%%%%%%%%%%%%%%%%%%%%%%%%%%%%%%%%%%%%%
%%%%%%%%%%%%%%%%%%%%%%%%%%%%%%%%%%%%%%%%%%%%%%%%%%%%%%%%%%%%%%%%
\vspace{-0.2cm}
\section{Conclusion}
\label{Sec_Conclusion}

We introduce SwitchMT, a novel methodology with adaptive task-switching strategy for RL-based multi-task learning that transitions between tasks based on the learning progress. 
By monitoring reward and internal dynamics of the model parameters, SwitchMT determines effective switching points, reducing training time and preventing overfitting. 
This dynamic adjustment enhances learning efficiency, simplifies the training process, and reduces the time and energy for hyperparameter tuning. 
Experimental results show that our SwitchMT achieves competitive scores across multiple Atari games (i.e., Pong: -8.8, Breakout: 5.6, and Enduro: 355.2) and obtains higher game points as compared to the state-of-the-art, without increasing network complexity. 
These results demonstrate the potential of SwitchMT methodology to enable intelligent autonomous agents capable of performing simultaneous multi-task learning.

%%%%%%%%%%%%%%%%%%%%%%%%%%%%%%%%%%%%%%%%%%%%%%%%%%%%%%%%%%%%%%%%%%%%%%%%%%%%%%
%%%%%%%%%%%%%%%%%%%%%%%%%%%%%%%%%%%%%%%%%%%%%%%%%%%%%%%%%%%%%%%%%%%%%%%%%%%%%%
%% The acknowledgments section is defined using the "acks" environment
%% (and NOT an unnumbered section). This ensures the proper
%% identification of the section in the article metadata, and the
%% consistent spelling of the heading.
\vspace{-0.2cm}
\begin{acks}
This work was partially supported by the NYUAD Center for Artificial Intelligence and Robotics (CAIR), funded by Tamkeen under the NYUAD Research Institute Award CG010. 
\end{acks}

%%%%%%%%%%%%%%%%%%%%%%%%%%%%%%%%%%%%%%%%%%%%%%%%%%%%%%%%%%%%%%%%%%%%%%%%%%%%%%
%%%%%%%%%%%%%%%%%%%%%%%%%%%%%%%%%%%%%%%%%%%%%%%%%%%%%%%%%%%%%%%%%%%%%%%%%%%%%%
%% The next two lines define the bibliography style to be used, and
%% the bibliography file.
\bibliographystyle{ACM-Reference-Format}
\bibliography{bibliography}
\end{spacing}

%% If your work has an appendix, this is the place to put it.
% \appendix

% \section{Research Methods}

% \subsection{Part One}

% Lorem ipsum dolor sit amet, consectetur adipiscing elit. Morbi
% malesuada, quam in pulvinar varius, metus nunc fermentum urna, id
% sollicitudin purus odio sit amet enim. Aliquam ullamcorper eu ipsum
% vel mollis. Curabitur quis dictum nisl. Phasellus vel semper risus, et
% lacinia dolor. Integer ultricies commodo sem nec semper.

% \subsection{Part Two}

% Etiam commodo feugiat nisl pulvinar pellentesque. Etiam auctor sodales
% ligula, non varius nibh pulvinar semper. Suspendisse nec lectus non
% ipsum convallis congue hendrerit vitae sapien. Donec at laoreet
% eros. Vivamus non purus placerat, scelerisque diam eu, cursus
% ante. Etiam aliquam tortor auctor efficitur mattis.

% \section{Online Resources}

% Nam id fermentum dui. Suspendisse sagittis tortor a nulla mollis, in
% pulvinar ex pretium. Sed interdum orci quis metus euismod, et sagittis
% enim maximus. Vestibulum gravida massa ut felis suscipit
% congue. Quisque mattis elit a risus ultrices commodo venenatis eget
% dui. Etiam sagittis eleifend elementum.

% Nam interdum magna at lectus dignissim, ac dignissim lorem
% rhoncus. Maecenas eu arcu ac neque placerat aliquam. Nunc pulvinar
% massa et mattis lacinia.

\end{document}